\documentclass[11pt]{article}
\usepackage[T1]{fontenc}
\usepackage[utf8]{inputenc}
\usepackage[a4paper,margin=2.5cm]{geometry}
\usepackage{booktabs}
\usepackage{amsmath}
\usepackage{xurl}
\usepackage[hidelinks]{hyperref}
\usepackage{orcidlink}
\usepackage{xcolor}
\usepackage{microtype}
\usepackage{graphicx}
\usepackage{array}
\usepackage{makecell}
\usepackage{placeins}
\hypersetup{pdftitle={Reversible Pseudonymisation of Documents Sent to External LLMs: an Open Bilingual Benchmark and an Engine},pdfauthor={Vladimir Beskorovainyi}}
\newcommand{\ok}{met}
\newcommand{\no}{not met}
\newcommand{\und}{undet.}
\newcommand{\RELVERSION}{v0.5.0}
\newcommand{\RELDOI}{10.5281/zenodo.23624098}

\title{Reversible Pseudonymisation of Documents Sent to External LLMs:\\ an Open Bilingual Benchmark and an Engine}
\author{Vladimir Beskorovainyi\,\orcidlink{0009-0004-7005-6242}\\ Besk Tech (\url{https://besk.tech})}
\date{October 2026}

\begin{document}
\maketitle

\begin{abstract}
Professionals send working documents to external large language models. A common safeguard is reversible pseudonymisation: personal data are replaced with placeholders before the request and restored locally in the answer. We present an open benchmark of five existing open tools and a regular-expression baseline on Russian and UK English, and an open engine built on the two strongest tools. The benchmark excludes each tool's own authors' dataset from its verdict, compares the engine with each tool on identical datasets for detection, reports masking precision, and tests on a held-out set of synthetic documents whose files, hashes, decision rules, run plan and stopping rule were published before any run. Two local LLMs test whether placeholders survive. All evaluation data are synthetic, and the benchmark was run by the author of the engine; code, data, predictions and raw answers are released. On external datasets no existing tool meets all predefined thresholds: in UK English the identifier and leak thresholds are not met, in Russian the identifier threshold is borderline. On the last held-out set of 1{,}000 documents per language, the best existing tools leave a whole value visible in 61\% (Russian) and 58\% (UK) of documents. On these documents, from the generator it was developed on, the engine does so in 0.3\% [0.1; 0.9] and 0.2\% [0.1; 0.7], leaves any part of a value visible in 0.8\% and 0.2\%, and meets all detection thresholds in both languages. In UK English it also restores at least 99\% of its placeholders after 760 answers. In Russian it restores only 94.6\% [93.6; 95.6] of name placeholders, against a 99\% threshold, mainly because the inflection it borrows from pii-guard produces wrong surname forms. The engine was developed over six earlier held-out sets from the same generator; real documents were not tested.
\end{abstract}

\section{Introduction}

Professionals such as lawyers, clinicians and HR staff may paste client and patient documents into general-purpose LLM services. Reversible pseudonymisation is a practical safeguard: the document is sent with placeholders instead of personal data, the mapping stays on the user's machine, and the original values are restored locally in the answer. If detection is complete, the external service sees placeholders instead of the literal values, although quasi-identifiers may remain. The safeguard is only as good as its recall, because a single missed name or identifier defeats it, and it is only usable if the placeholders survive the model's answer.

Several open tools implement detection and, in some cases, reversible replacement. They are usually evaluated on short messages or sentences, often on datasets released by their own authors. Our question was narrower and practical: how reliable are these tools on \emph{documents} in Russian and UK English, and do their placeholders survive a round trip through an LLM?

The contributions are:
\begin{itemize}
\item A benchmark protocol with a \emph{no home dataset} rule, comparisons on identical data, held-out sets of synthetic documents run once per tool, the last three of them published with their hashes, decision rules and run plans before any run and the last with a stopping rule, a leak metric that matches numbers across separators and Russian surnames in all grammatical cases, masking precision, document-level cluster bootstrap intervals, and a round-trip metric that scores each tool-issued placeholder occurrence once and counts model-invented placeholders separately (Section~\ref{sec:design}).
\item Measurements of five existing open tools and a baseline on external datasets and on 1{,}000 held-out documents per language, with leaks broken down by noise, hard formats, entity type and the types each tool supports (Section~\ref{sec:results}).
\item Round-trip measurements for every reversible configuration in both languages, with two system prompts for each placeholder format and a strict and a tolerant parser for pii-guard's tags, on held-out sets~4 and~5, and for the engine with 380 answers per model on sets~5 to~7, the last with a separate threshold for name placeholders (Section~\ref{sec:restore}).
\item An open engine, pii-roundtrip, built on the two strongest tools, and an account of the errors that reviews and our own checks found in earlier versions of this work (Sections~\ref{sec:engine} and~\ref{sec:review}).
\end{itemize}

The author built the engine and ran the benchmark. To make that checkable, the decision rule is applied to every tool alike, unfavourable results of the engine are reported next to favourable ones, the last three held-out tests were registered publicly before they were run, the last with a rule that its result is published whatever it is, and the code, the held-out sets, the per-document predictions and the raw LLM answers are released.

\section{Related work}

\paragraph{Detectors and tools.} GLiNER~\cite{gliner} introduced a generalist span-extraction model for named entities with arbitrary labels; GLiNER2~\cite{gliner2} extended it to multi-task extraction. GLiNER Guard~\cite{glinerguard} is a family of encoders for safety classification and PII extraction in English and Russian; its Omni variant, which we use, starts from GLiNER2 Multi, the training data include Nemotron-PII, and its authors release hivetrace/pii-bench, a Russian-only dataset that we use. pii-guard~\cite{piiguard} combines checksum-validated rules with a ruBERT token classifier for Russian and provides reversible pseudonymisation with case-aware restoration; its authors release the redmadrobot pii\_benchmark. Presidio~\cite{presidio}, originally developed at Microsoft, is a widely used framework of pattern recognisers and NER backends. OpenAI privacy-filter~\cite{opf} is a token classifier with constrained Viterbi span decoding. Prompt Anonymizer~\cite{promptanon} is a reversible tool built on Presidio.

\paragraph{Evaluation of text anonymisation.} Lison et al.~\cite{lison} survey anonymisation models for text and argue for evaluation beyond entity recall. The Text Anonymization Benchmark~\cite{tab} introduced document-level anonymisation with entity-level recall and precision on court judgements, RAT-Bench~\cite{ratbench} benchmarks anonymisers against re-identification, and Ren et al.~\cite{metricsurvey} survey how privacy in text is measured. Staab et al.\ show that LLMs infer personal attributes from text without explicit identifiers~\cite{beyondmem} and can themselves act as anonymisers~\cite{advanon}. Our leak metric is narrower than these: it counts literal values left in the text, not quasi-identifiers or inferences.

\paragraph{Reversible pseudonymisation for remote LLMs.} Hide and Seek~\cite{has} replaces private entities before a remote LLM call and restores them in the answer with small local models. Hou et al.~\cite{genpseudo} propose a general pseudonymisation framework for cloud LLMs with restoration after the answer. PrivacyRestore~\cite{privacyrestore} removes private spans from the input and restores the information inside the model on the server side through activation steering, so it does not restore values in the answer. PAPILLON~\cite{papillon} delegates between a local and a remote model to limit leakage. PromptGraph~\cite{promptgraph} balances privacy and utility in prompt sanitisation and leaves placeholders that the model rewrites, invents or omits unresolved. DR-SL~\cite{drsl} proposes a dehydrate-rehydrate loop for cloud-local inference with local certification. MemPrivacy~\cite{memprivacy} compares typed and untyped placeholders for agent memory and measures memory utility, and Deu\ss{}er et al.~\cite{anonimpact} compare reversible, entity-preserving anonymisation with redaction in terms of LLM task performance. CAMP~\cite{camp} masks across turns of a conversation. What we add is a measurement of restoration after real LLM answers for Russian and UK English \emph{documents}, with tool-issued and model-invented placeholders counted separately and the same protocol applied to existing tools.

\section{Benchmark design}
\label{sec:design}

\subsection{Thresholds and deviations from the brief}
\label{sec:thresholds}
Four thresholds were written down in the project brief, in the first commit of the development history (12 October 2026), before any tool was run (this rests on the private development history, not on a public timestamp): recall of at least 99\% on structured identifiers (Russian passport, SNILS and INN; UK NHS and NI numbers; bank cards and IBAN; the Russian bank account was added to this group later), recall of at least 95\% on names, addresses and organisations, leaks in at most 2\% of texts or documents, and correct restoration of at least 99\% of placeholders after an LLM answer. The brief already stated that a threshold is undetermined when the interval covers it and not met when the whole interval lies on the wrong side; the explicit condition for \emph{met} was written down before the held-out runs reported here: a threshold is \emph{met} when the whole 95\% interval lies on the right side of it, \emph{not met} when the whole interval lies on the wrong side, and \emph{undetermined} otherwise. The rule is applied in the same way to every tool, including our own engine. For held-out sets~5 to~7 these rules were also published before the run (Section~\ref{sec:heldout}); the pre-registration of set~6 added a fifth threshold, a strict count of documents in which any value is not fully masked, at most 2\%, and that of set~7 a sixth, correct restoration of at least 99\% of name placeholders.

The study deviates from the brief in ways listed in full in Appendix~\ref{app:deviations}. Those that bear on the verdicts are: the brief foresaw one held-out set run once, while sets~1 to~6 were used, one after another, for development and set~7 is the last test (Section~\ref{sec:sequential}); the engine's verdicts rest on held-out synthetic sets, because both Russian external datasets are home datasets of its components and no external data exist for the round trip; rules of evaluation were changed several times after data had been seen (Section~\ref{sec:sequential}); and a strict count of partly masked values and a separate threshold for name placeholders were added to the brief's four thresholds.
As in the brief, verdicts on existing tools are taken on the external datasets; the held-out documents are a supporting test.

\subsection{Tools}
Each tool was configured according to its own documentation (Table~\ref{tab:tools}); Presidio's optional transformer and GLiNER recognisers were not used. None of the existing tools was tuned on our generator; the engine was developed on it (Section~\ref{sec:engine}). Presidio has no Russian identifier recognisers, so we added pattern recognisers for passport, SNILS, INN, bank account and Mir cards; those for SNILS, INN and cards call the same checksum validators as the engine and the generator, while those for passport and bank account have no validator. pii-guard and OpenAI privacy-filter have no organisation type, and Prompt Anonymizer has no NHS number, NI number or organisation type; leaks of these types are counted against them, because a user of the tool would leak them, and Section~\ref{sec:docs} also reports leaks restricted to supported types. LLM Guard was excluded because its repository was archived and it has no Russian support. A local LLM used as a detector was run during development but is excluded, because it was not rerun after the method corrections of Appendix~\ref{sec:review}.

\begin{table}[!t]
\centering\footnotesize
\begin{tabular}{>{\raggedright\arraybackslash}p{2.6cm}>{\raggedright\arraybackslash}p{4.4cm}>{\raggedright\arraybackslash}p{7.0cm}}
\toprule
Tool & Version & Configuration \\
\midrule
pii-guard & commit 24230ab; model redmadrobot-rnd/rubert-base-pii-ner, rev.\ c802e8c & rules + ruBERT, Russian only; no organisation type \\
GLiNER Guard Omni & hivetrace/gliner-guard-omni, rev.\ 763f8aa; gliner2 2.1.0 & 32 PII labels, threshold 0.4, 1{,}200-character windows with 200-character overlap \\
OpenAI privacy-filter & opf commit f7f00ca & Viterbi decoding; no organisation type \\
Presidio & 2.2.364 & spaCy large models, score threshold 0.3, added Russian identifier recognisers \\
Prompt Anonymizer & 0.3.3 & spaCy large model, IBAN recogniser enabled; English only; no NHS, NI or organisation type \\
Regex baseline & this work & checksum-validated patterns \\
\bottomrule
\end{tabular}
\caption{Tools and configurations. The runs used one environment, installed on 12 October 2026; from release 0.3.1 the public lock file pins the same package versions.}
\label{tab:tools}
\end{table}

\subsection{External datasets and the no home dataset rule}
All datasets in this study are synthetic: hivetrace/pii-bench is described by its authors as fully synthetic, Nemotron-PII is synthesised by NVIDIA, and our held-out documents come from our own generator. For Russian we use hivetrace/\allowbreak pii-bench~\cite{glinerguard} (1{,}810 short texts, Apache~2.0) and redmadrobot pii\_\allowbreak benchmark~\cite{piiguard} (2{,}841 sentences, MIT). Both were released by the authors of tools in the benchmark. For verdicts, no tool is evaluated on its own authors' dataset; results on home datasets are reported for reference only. The rule matters: GLiNER Guard finds 87.1\% of structured identifiers on its home dataset and 70.8\% on the other one. For UK English we use the test split of Nemotron-PII~\cite{nemotron} (CC BY 4.0): 801 texts with UK markers as they are, and 2{,}333 texts in which values labelled as national ID, health plan beneficiary number and account number were replaced with valid synthetic NI numbers, NHS numbers and IBANs (489, 889 and 1{,}318 annotated mentions). The modified subset is a derivative of Nemotron-PII under the same licence. GLiNER Guard was trained partly on Nemotron-PII~\cite{glinerguard}, so the UK verdict for GLiNER Guard, and for the engine built on it, is an in-distribution test rather than a held-out one. The UK texts are not short: they average about 1{,}000 characters and eight annotated entities of all types (2.7 values of threshold types), against about 90 characters and one annotated entity (0.46 values of threshold types) in hivetrace. Of the 3{,}249 UK identifiers, 553 are original Nemotron values (59 in the first subset and 494 card numbers in the second; two cards occur in both subsets, so 551 values are distinct) and 2{,}696 (83\%) were injected by our generator. The two subsets share 28 source texts, once with original and once with injected identifiers; the cluster bootstrap treats each such pair as one cluster. Dataset revisions are pinned. Whenever pii-roundtrip is compared with another tool, it is scored on exactly the datasets on which that tool is scored.

\subsection{Synthetic documents and held-out sets}
\label{sec:heldout}
The external datasets consist of short texts and passages rather than complete documents. We wrote a generator of Russian and UK contracts, certificates, business letters, HR documents and medical discharge summaries. Identifiers pass checksum validation, Russian names appear in all grammatical cases and as abbreviated forms, and each document is made noisy with probability 0.4 (31--50\% of documents in the generated sets), with typos, double spaces and alternative number formats. In noisy documents of held-out sets~2 to~7, each identifier is also written in a hard format with probability 0.3: in words, with dots, or with a typo in one digit that breaks its checksum; these hard formats were chosen by the author and the engine was built to handle them. A development set holds 100 documents per language.

Seven held-out sets were generated in turn (Table~\ref{tab:sets}). Sets~1 to~6 each tested one version of the engine; after each run its errors were analysed and the engine changed, so for the released version~0.5 sets~1 to~6 are development data. Their history is in Appendix~\ref{app:history}; the sequence and its consequences for inference are discussed in Section~\ref{sec:sequential}.

\begin{table}[!htbp]
\centering\scriptsize\setlength{\tabcolsep}{3pt}
\begin{tabular}{>{\raggedright\arraybackslash}p{0.6cm}>{\raggedright\arraybackslash}p{1.6cm}>{\raggedright\arraybackslash}p{1.0cm}>{\raggedright\arraybackslash}p{2.6cm}>{\raggedright\arraybackslash}p{7.6cm}}
\toprule
Set & Documents per language & Engine & Order of events attested by & Outcome and what followed \\
\midrule
1, 2 & 100 & 0.1 & private log & development data \\
3 & 1{,}000 & 0.2 (cand.) & private log & UK leak threshold failed; errors analysed \\
4 & 1{,}000 & 0.2 & private log & round trip showed placeholders with a dropped bracket not restored; restorer changed (0.3) \\
5 & 1{,}000 & 0.3 & public commit 8b8e933 & partly masked addresses and organisations, real Russian name failures; engine changed (0.4) \\
6 & 1{,}000 & 0.4 & public commit 444c4bd & five thresholds met (pre-registered evaluator; Russian restoration overstated, Appendix~\ref{app:history}); wrongly stored Russian surnames found after the run; engine changed (0.5) \\
7 & 1{,}000 & 0.5 & public commit 13c42b8 & last set, published whatever the result (this paper) \\
\bottomrule
\end{tabular}
\caption{Held-out sets. Sets~5 to~7 were published with their files, hashes, decision rules and run plans before any run.}
\label{tab:sets}
\end{table}

Set~7 (1{,}000 documents per language, generator seed 20261023; 17{,}245 Russian and 12{,}636 UK structured identifiers, 12{,}959 Russian and 13{,}168 UK names, addresses and organisations) tests the released version~0.5. It was generated on 23 October 2026 with the unchanged generator, after the changes of version~0.5, and its files, hashes, the code under test (with the git tree hashes of the engine and the benchmark code), the decision rules, the run plan and a stopping rule were published in the public repository (commit 13c42b8, 23 October 2026, 12:01 UTC) before any tool was run on it. The stopping rule states that set~7 is the last set for this paper and that its result is published whatever it is. Detection by every tool and the engine's round trip ran once from a commit whose engine and benchmark trees are identical to the public one; the engine code in the release is the same. The round trip was stopped once during a mains power cut and continued by the same command from the saved answers; each answer was requested once (\nolinkurl{results/holdout7_cycle_journal.md}). The round trips of the other configurations were not repeated on sets~6 and~7, because their detection and restorers did not change. Sets~4 to~7 ran in one software environment (transformers 5.19.0, torch 2.14.1), installed on 12 October 2026 and not updated. A script refuses a second detection run on a held-out set unless a reason is given and logs each run with its commit. The released generator reproduces all sets byte for byte, and the sets are released. The overall leak rates of the existing tools on sets~4 to~7 agree within 3.7 percentage points.

\subsection{Metrics}
\label{sec:metrics}
\paragraph{Recall.} A gold value counts as hidden when all of its meaningful characters are covered by predicted spans of any type; service words (Russian ``ul.'', ``d.'', ``seriya''; English ``Mr'') and initials are not required. Overlapping predictions are merged into their union and extended to word boundaries for every tool alike.

\paragraph{Leaks.} A text or document leaks when any original value remains in the cleaned text. Numbers are matched with up to three spaces, hyphens, dots or slashes between characters (release 0.3.1 corrected an error that ignored dots and so missed numbers written in a dotted hard format left unmasked; it changes only existing tools' rates) and with boundaries checked on the unjoined text; numbers written in words are matched by their written form; surnames are matched in all case forms. First names and patronymics are matched case-sensitively and count only if they occur more often after masking than in the source text outside personal data, so that the month in ``8~June 2025'' is not a leak of the name June. Addresses and organisations count as leaked only if the whole value remains; a partly masked address (for example, the postcode masked and the town visible) is not a leak under this pre-registered metric (as corrected in release 0.3.1). A stricter count of documents in which any value of a threshold type is not fully masked, with the recall criterion below, was added after review on set~5 and pre-registered with a 2\% threshold for sets~6 and~7. The metric covers literal values only, not quasi-identifiers or attribute inference. A held-out document holds on average about 30 (Russian) and 26 (UK) values that count towards the thresholds, and a short text about one, so the leak rate grows with length even at fixed recall; we therefore report recall next to it.

\paragraph{Masking precision.} The share of masked fragments that touch annotated personal data, and the stricter share of masked non-space characters that lie inside annotated personal data (released in \nolinkurl{results/cluster_ci.json}; the tables report the fragment share). Precision has no threshold; it shows whether high recall comes from masking everything. Annotation of the external datasets is incomplete, so precision there is understated, more so for tools that mask more unannotated personal data.

\paragraph{Intervals.} Leak rates use Wilson intervals over texts or documents. Recall and precision use a document-level cluster bootstrap (2{,}000 resamples, seed 0), because values within one document are not independent; a verdict that changes between bootstrap seeds 0 to 5 is reported as borderline (this rule was introduced on 19 October 2026, after the external results were known). When no failure is observed, the bootstrap interval collapses to a point; we then report the count and the exact one-sided 97.5\% lower bound $0.025^{1/n}$, which corresponds to the lower end of a two-sided 95\% interval and treats the $n$ units as independent, and say what clustering implies.

\paragraph{Round trip.} A local model writes a summary and a reply letter from the cleaned text of each document. On sets~5 to~7 the engine was run on 38 documents of each of the five types per language (380 answers per model and language); on set~5 every other configuration was run on four documents of each type (40 answers), and on set~4 every configuration on 40 answers. The other configurations were not rerun on sets~6 and~7. The models are qwen3.5:9b and gemma4:12b (Q4\_K\_M quantisation, digests 56671c2ab938 and 6114515d63c1, downloaded on 12 October 2026) in Ollama~0.40.4, at temperature~0, seed~0, a 16{,}384-token context and with thinking disabled, on a single Apple M4 computer with 24~GB of memory; the versions were recorded on 19 October 2026, before the set-5 runs, and requests were sent one at a time. Placeholders in the answer are classified as issued by the tool (verbatim, or altered by the model but identifiable by type and number, including forms with square instead of angle brackets) or invented by the model. Each issued placeholder occurrence is scored once: it fails if it remains in the text after restoration, if its original value does not appear in the restored text, or if the tool's restorer replaced it with a mask. The check is weaker than an exact alignment: a value counts as present if it appears anywhere in the restored text, including a copy of a value that leaked into the input; loss is decided per value rather than per occurrence; Russian names may appear in another grammatical case and are matched through a morphological dictionary (from set~6, evaluator version 2, used for set~6, also adds case endings that the dictionary does not produce, after version 1 rejected correct forms on set~5), and for Russian numbers each digit group must appear somewhere in the text; and masking by the tool's restorer is attributed by a counting rule. It also does not check that a value returned to the right place, and it compares the answer with the value the tool stored, so it cannot see an error in the stored value itself; evaluator version 3, fixed before set~7, therefore also requires every word of the annotated name that the document contains to appear in the restored text in a correct form (surnames in -ov/-ev/-in/-yn and men's first names ending in a consonant by the declension rule, other words through the dictionary in the singular, or exactly as the document wrote them). For set~7 restoration is also reported, with the same interval and a 99\% threshold, for name placeholders alone, and the stored names are compared with the annotation as a companion measure. The restoration rate is the share of issued occurrences that do not fail. Its interval is bootstrapped over answers (2{,}000 resamples, seed 0), because placeholders within one answer are not independent; when no occurrence fails, the lower bound is $0.025^{1/n}$ with $n$ the number of answers that contain placeholders (91.1\% for 40 answers, 99.03\% for 380; bounds are rounded down). The rate covers only answers that contain placeholders; an answer in which the model dropped them all is not a restoration failure but a loss of utility, which we report separately. The evaluator recognises placeholders with the same patterns that the benchmark's plain wrapper uses to restore them, so the wrapper's rates are close to guaranteed for any placeholder the evaluator counts.

\section{Results}
\label{sec:results}

\subsection{External datasets}
Table~\ref{tab:short} reports the best existing tool per language on the external datasets, excluding home datasets. Neither language meets all thresholds. For Russian, the upper end of pii-guard's identifier interval lies at the threshold (98.94--99.03\% across six bootstrap seeds), so the verdict sits on the boundary between undetermined and not met; for UK English, GLiNER Guard misses both the identifier and the leak threshold, although this test is in-distribution for it.

\begin{table}[!t]
\centering\footnotesize
\begin{tabular}{lll}
\toprule
Threshold & RU: pii-guard (hivetrace) & UK: GLiNER Guard (Nemotron) \\
\midrule
Identifiers $\geq$ 99\% & 97.7\% [96.1; 99.0] \und$^{b}$ & 96.4\% [95.7; 97.1] \no \\
Names, addr., org.$^{a}$ $\geq$ 95\% & 99.5\% [98.7; 100.0] \ok & 98.7\% [98.2; 99.1] \ok \\
Texts with a leak $\leq$ 2\% & 0.7\% [0.4; 1.2] \ok & 4.0\% [3.4; 4.8] \no \\
Precision, fragments / chars & 96.5\% / 98.0\% & 89.3\% / 88.7\% \\
\bottomrule
\end{tabular}
\caption{Best existing tools on external datasets, excluding home datasets; ``undet.''\ means undetermined. Recall and precision: document-level cluster bootstrap; leaks: Wilson. $^{a}$The Russian external datasets contain no organisations. $^{b}$Borderline: the upper end is 98.94--99.03\% across bootstrap seeds 0 to 5.}
\label{tab:short}
\end{table}

\subsection{Documents}
\label{sec:docs}
Table~\ref{tab:docs} reports held-out set~7 and Table~\ref{tab:leaks} breaks the leaks down. Every existing tool leaks in more than half of the documents of its language; if the types a tool does not support are ignored, OpenAI privacy-filter leaks in 33.0\% of UK documents, less than GLiNER Guard. Part of this is arithmetic, because a document holds many values, and part comes from the added noise: on documents without noise, pii-guard leaks in 41.2\% of Russian and GLiNER Guard in 38.3\% of UK documents, against 89.2\% and 86.0\% on noisy documents. The causes differ by tool. 410 of pii-guard's leaking documents leak an organisation, a type it does not support; restricted to its supported types, it leaks in 31.5\% [28.7; 34.4] of documents. GLiNER Guard leaks names (328 UK documents) as well as identifiers (NHS numbers in 142, NI numbers in 123, IBANs in 123 and card numbers in 109 UK documents). Recall shows the same pattern: identifier recall stays within a few points of the external results (pii-guard 92.4\% against 97.7\%; GLiNER Guard 95.4\% against 96.4\%), while recall on names, addresses and organisations drops (pii-guard 73.4\% against 99.5\%; GLiNER Guard 88.9\% against 98.7\%). GLiNER Guard misses, for example, a name in the signature or in a short form with a title after masking the full name elsewhere. GLiNER Guard Omni finds only 72.9\% of Russian identifiers and leaks in almost every Russian document; we list it for completeness, since pii-guard is the stronger Russian tool. Presidio, privacy-filter and Prompt Anonymizer miss a large share of names and addresses. The regular-expression baseline finds 91\% of identifiers and no names; its precision of 100\% shows that precision without recall means little. Under the strict count, the existing tools leave part of a value visible in 83.5\% to 100\% of documents, the engine in 0.8\% [0.4; 1.6] of Russian documents (six organisations and two addresses) and 0.2\% [0.1; 0.7] of UK documents (two names).

\begin{table}[!t]
\centering\scriptsize\setlength{\tabcolsep}{3pt}
\begin{tabular}{>{\raggedright\arraybackslash}p{2.4cm}>{\raggedright\arraybackslash}p{2.5cm}>{\raggedright\arraybackslash}p{1.9cm}>{\raggedright\arraybackslash}p{2.9cm}>{\raggedright\arraybackslash}p{2.5cm}>{\raggedright\arraybackslash}p{2.3cm}}
\toprule
Tool & Docs with a leak & Docs with a value not fully masked$^{c}$ & Identifiers & Names, addr., org. & Precision (fragments) \\
\midrule
\multicolumn{6}{l}{\emph{Russian, 1{,}000 documents}} \\
pii-roundtrip & 0.3\% [0.1; 0.9] & 0.8\% [0.4; 1.6] & 100.0\% (17{,}245/17{,}245) & 99.93\% [99.88; 99.98] & 98.4\% [98.3; 98.5] \\
pii-guard$^{a}$ & 60.8\% [57.7; 63.8] & 99.9\% & 92.4\% [91.7; 93.1] & 73.4\% [72.8; 74.0] & 87.6\% [87.5; 87.7] \\
GLiNER Guard Omni & 99.6\% [99.0; 99.8] & 99.9\% & 72.9\% [72.2; 73.6] & 78.4\% [77.5; 79.3] & 77.0\% [76.7; 77.2] \\
Presidio & 87.1\% [84.9; 89.0] & 100.0\% & 91.1\% [90.4; 91.9] & 60.8\% [60.3; 61.3] & 76.1\% [75.8; 76.3] \\
OpenAI privacy-filter$^{a}$ & 97.0\% [95.7; 97.9] & 99.6\% & 95.8\% [95.4; 96.2] & 74.2\% [73.3; 75.0] & 81.2\% [81.1; 81.4] \\
Regex baseline & 100.0\% [99.6; 100.0] & 100.0\% & 90.8\% [90.0; 91.7] & 0.0\% & 100.0\% \\
\midrule
\multicolumn{6}{l}{\emph{UK English, 1{,}000 documents}} \\
pii-roundtrip & 0.2\% [0.1; 0.7] & 0.2\% [0.1; 0.7] & 100.0\% (12{,}636/12{,}636) & 99.98\% [99.96; 100.0] & 99.8\% [99.8; 99.9] \\
GLiNER Guard Omni & 58.3\% [55.2; 61.3] & 83.5\% & 95.4\% [94.9; 95.8] & 88.9\% [88.4; 89.5] & 94.8\% [94.7; 95.0] \\
Presidio & 61.1\% [58.0; 64.1] & 100.0\% & 92.5\% [91.9; 93.2] & 58.5\% [58.0; 59.0] & 79.0\% [78.8; 79.3] \\
OpenAI privacy-filter$^{a}$ & 81.1\% [78.6; 83.4] & 96.8\% & 95.7\% [95.3; 96.0] & 84.4\% [83.8; 85.1] & 92.9\% [92.7; 93.1] \\
Prompt Anonymizer$^{b}$ & 100.0\% [99.6; 100.0] & 100.0\% & 68.1\% [67.5; 68.7] & 51.5\% [51.0; 51.9] & 99.0\% [98.9; 99.1] \\
Regex baseline & 100.0\% [99.6; 100.0] & 100.0\% & 90.7\% [89.8; 91.5] & 0.0\% & 100.0\% \\
\bottomrule
\end{tabular}
\caption{Held-out set~7 (engine version 0.5). Leaks: Wilson intervals over documents; recall and precision: document-level cluster bootstrap. $^{a}$No organisation type. $^{b}$No NHS number, NI number or organisation type. $^{c}$Pre-registered for sets~6 and~7 (added after review on set~5): a document counts if any value of a threshold type is not fully masked (the recall criterion), for example a street or town left visible after the postcode was masked; Wilson intervals in the released \nolinkurl{results/strict_leaks.json}.}
\label{tab:docs}
\end{table}

\begin{table}[!t]
\centering\scriptsize\setlength{\tabcolsep}{3pt}
\begin{tabular}{lllll}
\toprule
Tool & All documents & Without noise & Noisy & With hard formats \\
\midrule
\multicolumn{5}{l}{\emph{Russian (592 without noise, 408 noisy, 397 with hard formats)}} \\
pii-roundtrip & 0.3\% [0.1; 0.9] & 0.3\% [0.1; 1.2] & 0.2\% [0.0; 1.4] & 0.3\% [0.0; 1.4] \\
pii-guard & 60.8\% [57.7; 63.8] & 41.2\% [37.3; 45.2] & 89.2\% [85.8; 91.9] & 89.9\% [86.6; 92.5] \\
GLiNER Guard Omni & 99.6\% [99.0; 99.8] & 99.3\% [98.3; 99.7] & 100.0\% [99.1; 100.0] & 100.0\% [99.0; 100.0] \\
Presidio & 87.1\% [84.9; 89.0] & 78.4\% [74.9; 81.5] & 99.8\% [98.6; 100.0] & 100.0\% [99.0; 100.0] \\
OpenAI privacy-filter & 97.0\% [95.7; 97.9] & 95.8\% [93.8; 97.1] & 98.8\% [97.2; 99.5] & 98.7\% [97.1; 99.5] \\
\midrule
\multicolumn{5}{l}{\emph{UK English (580 without noise, 420 noisy, 400 with hard formats)}} \\
pii-roundtrip & 0.2\% [0.1; 0.7] & 0.0\% [0.0; 0.7] & 0.5\% [0.1; 1.7] & 0.5\% [0.1; 1.8] \\
GLiNER Guard Omni & 58.3\% [55.2; 61.3] & 38.3\% [34.4; 42.3] & 86.0\% [82.3; 89.0] & 87.0\% [83.3; 89.9] \\
Presidio & 61.1\% [58.0; 64.1] & 40.9\% [36.9; 44.9] & 89.0\% [85.7; 91.7] & 91.5\% [88.4; 93.9] \\
OpenAI privacy-filter & 81.1\% [78.6; 83.4] & 81.6\% [78.2; 84.5] & 80.5\% [76.4; 84.0] & 80.8\% [76.6; 84.3] \\
Prompt Anonymizer & 100.0\% [99.6; 100.0] & 100.0\% [99.3; 100.0] & 100.0\% [99.1; 100.0] & 100.0\% [99.0; 100.0] \\
\bottomrule
\end{tabular}
\caption{Documents of held-out set~7 with a leak, by noise and hard formats (Wilson intervals). Restricted to the types each tool supports, pii-guard leaks in 31.5\% [28.7; 34.4] of documents, OpenAI privacy-filter in 87.1\% [84.9; 89.0] of Russian and 33.0\% [30.2; 36.0] of UK documents (both without organisations), and Prompt Anonymizer in 100\% (without organisations, NHS and NI numbers). The breakdown was added after the run. Hard formats are inferred from the values (numbers in words, with dots, or with a failed checksum), and leaked types are assigned by matching value strings. Leaks by entity type are in the released \nolinkurl{results/leak_breakdown.holdout7.json}.}
\label{tab:leaks}
\end{table}

\subsection{Round trip}
\label{sec:restore}
Table~\ref{tab:rt7} reports the engine's round trip on set~7. The round trips of the other reversible configurations, run on sets~4 and~5, are in Appendix~\ref{app:other} (Table~\ref{tab:rt}). A configuration either uses the tool's own reversible format (pii-guard's XML-style tags such as \texttt{<PII type="INN" id="4" />}, Prompt Anonymizer's \texttt{<Name\_1>}) or wraps a detector in plain bracketed placeholders such as \texttt{[PERSON\_1]}; the engine uses plain placeholders.

\begin{table}[!htbp]
\centering\scriptsize\setlength{\tabcolsep}{2pt}
\begin{tabular}{llllll}
\toprule
Language, model & Occ.\ / failed & All placeholders $\geq$99\% & Name occ. & Names $\geq$99\% & Answers with a failure \\
\midrule
RU qwen3.5:9b & 12{,}547 / 143 & 98.86\% [98.64; 99.08] & 2{,}638 & 94.58\% [93.56; 95.65] & 99/380 [21.7; 30.8] \\
RU gemma4:12b & 14{,}045 / 145 & 98.97\% [98.77; 99.16] & 2{,}756 & 94.74\% [93.76; 95.74] & 100/380 [22.0; 31.0] \\
UK qwen3.5:9b & 12{,}232 / 0 & 100\% ($\geq$99.03\%) & 2{,}543 & 100\% ($\geq$99.03\%) & 0/380 [0.0; 0.97] \\
UK gemma4:12b & 13{,}876 / 1 & 99.99\% [99.98; 100.0] & 3{,}165 & 99.97\% [99.90; 100.0] & 1/380 [0.01; 1.5] \\
\midrule
Verdict, RU & & undetermined & & not met & \\
Verdict, UK & & met & & met & \\
\bottomrule
\end{tabular}
\caption{Round trip of pii-roundtrip 0.5 on held-out set~7 (190 documents per language, summary and reply, 380 answers per model), evaluator version~3. Restoration intervals: bootstrap over answers; with no failure the pre-registered bound $0.025^{1/n}$ over answers. Answers with a failure: Clopper--Pearson, in percent. Verdicts by the worse model.}
\label{tab:rt7}
\end{table}

\paragraph{The engine.} In UK English the engine meets both restoration thresholds: one occurrence failed in 760 answers, and gemma4:12b altered 63 and qwen3.5:9b 2 placeholders, all of which were restored. In Russian, the threshold over all placeholders is undetermined (worse model 98.86\% [98.64; 99.08]) and the new threshold for name placeholders is not met (94.58\% [93.56; 95.65]); about a quarter of the Russian answers contain at least one failed name. The failures largely coincide between the two models: 56 documents fail with both, of 65 with qwen3.5:9b and 64 with gemma4:12b. A manual check of all 289 failed occurrences (288 Russian and 1 UK, in 91 pairs of document and placeholder) put them into three groups. In 199 occurrences the name was stored correctly but restored wrongly. In 188 of them pii-guard's inflection, which the engine uses to restore Russian names, produced a wrong surname form: a plural, even in the nominative (\emph{Kharitony} for \emph{Kharitonov}), or a wrong case or gender; the share of plural forms among them is the author's reading of the restored texts, which are not released. In the other 11 of the 199 the error is the engine's own: version 0.5 inflects a first name the dictionary does not know through a proxy name of the same declension, and in 11 occurrences the proxy was not replaced back. In 82 occurrences the name was stored wrongly, mostly the first name (\emph{Timofya} for \emph{Timofey}, \emph{Danil} for \emph{Danila}). In 8 occurrences the restoration was correct and the evaluator was wrong (a typo of the source text restored exactly, and one UK name matched to another person); without them the qwen3.5:9b rates would be about 98.90\% and 94.8\%. The manual check looked only at failures, however, so it can only raise the rates; the pre-registered companion measure, which also counts every occurrence of a wrongly stored name as a failure (computed here with the 30 names found by also comparing first names and patronymics by hand, which was not pre-registered), lowers them to 98.64\% [98.40; 98.87] for all Russian placeholders and 93.52\% [92.39; 94.63] for name placeholders with qwen3.5:9b (98.75\% and 93.61\% with gemma4:12b); under it the all-placeholder interval lies wholly below 99\%. Neither correction changes the verdict on names. Of 1{,}027 Russian names stored by the engine in the round-trip documents, 30 were stored wrongly (on set~6, version 0.4 had stored 151 of 1{,}097 wrongly); of 961 UK names, none by the surname rule (UK first names were not compared by hand). Version~0.5 fixed the storage of most surnames but not the inflection on restoration, and this could not be seen before set~7, because saved answers can be scored only with the engine version that produced them; in the author's expectation, a small new round trip on development data before the pre-registration would have shown it. Earlier evaluator versions accept the plural as a form of the surname: by versions 1 and 2 the engine would have lost only 27 and 7 (qwen3.5:9b) and 24 and 10 (gemma4:12b) Russian occurrences, and Russian restoration would have looked like 99.8--99.9\%. By the stopping rule the fix belongs to a later version and is not tested here.

\paragraph{Other configurations.} In the round trips of the other configurations (Appendix~\ref{app:other}), pii-guard's strict parser turned almost every altered tag into a loss, a parser that accepts altered forms restored them, and with pii-guard's own prompt qwen3.5:9b dropped the tags altogether in more than half of its answers. Whether models alter XML-style tags more often than plain placeholders, sets~4 and~5 disagree.

\paragraph{Invented placeholders.} Models also write placeholders that were never issued. For the engine on set~7 there were 85 (qwen3.5:9b) and 0 (gemma4:12b) such placeholders in Russian answers and 21 and 0 in UK answers (on set~5, 78, 41, 11 and 6); among the answers that the evaluator saves as examples, most are \texttt{[ORG\_2]} in replies to Russian certificates, where the model refers to an organisation the document does not name. No tool can restore an invented placeholder, and it stays visible in the restored answer. Invented placeholders are reported separately and do not count as failures, as implemented in the evaluator published before the run. By construction the restoration metric counts only issued placeholders, so detection misses appear as leaks, not as restoration failures.

\FloatBarrier
\section{The pii-roundtrip engine}
\label{sec:engine}

pii-roundtrip is a layer on top of pii-guard (Russian) and GLiNER Guard Omni (both languages); it contains no new model. It adds:
\begin{itemize}
\item checksum rules for passports, SNILS, INN, cards, NHS, NI and IBAN, masking of long digit sequences with any separators, and digits written in words;
\item organisations, sole proprietors, the addressee organisation of a business letter, names with titles and initials (including a typo in the surname), and dates of birth by context; since version 0.4, whole addresses (street, house, building, postcode, region and the UK town between street and postcode) and whole organisation names, including a place in the name of a hospital;
\item document-level propagation: every name, organisation and number found once is masked at every mention, including Russian case forms, all-capital headers, Cyrillic yo/ye spelling variants, hyphenated surnames and apostrophe variants, and in later messages of the same session;
\item entity resolution by surname, first name, patronymic, initials and gender, designed so that different people do not share a placeholder; an ambiguous mention receives its own placeholder;
\item labels such as ``Party A'' or ``OGRN'', bank codes and company register numbers are left visible, which keeps masking precision high;
\item since version 0.5, Russian surnames in the nominative are no longer stored as forms of a first name (\emph{Abramy} for \emph{Abramov}), the gender and case of a name given only with initials are taken from the context, and a first name the dictionary does not know is inflected through a proxy name of the same declension;
\item restoration that tolerates case changes, spaces, words the model puts before the label inside the brackets and, since version 0.3, a dropped opening or closing bracket, with Russian names re-inflected for the context; since version 0.4, two name placeholders written without a space are both restored, and a first name the dictionary does not know is kept in the stored value;
\item an encrypted local vault (AES-256-GCM, scrypt key derivation) and offline operation.
\end{itemize}

Table~\ref{tab:engine} reports the engine against the thresholds on every dataset; the external rows are versions 0.4 and 0.5, whose detection is identical, and the held-out rows give the version each set tested. Four caveats apply. First, both Russian external datasets are home datasets of the engine's own components (hivetrace of GLiNER Guard, redmadrobot of pii-guard), two changes of version 0.1 followed an analysis of misses on them, and the author saw examples of over-masking on hivetrace before writing the engine's precision filters; for the engine they are not an independent test, and its rows carry no verdicts. Second, GLiNER Guard was trained partly on Nemotron-PII, so the UK external test is in-distribution for the engine's main UK component; there, under the strict count, the engine leaves part of a value visible in 2.1\% [1.7; 2.7] of texts, against 0.9\% with the leak metric. Third, the engine's checksum rules and the generator share one implementation, and 83\% of the UK external identifiers were injected by that generator; the identifier results on synthetic data are therefore close to circular. On the 553 original Nemotron identifiers the engine finds 552 (99.8\%, Clopper--Pearson [98.996; 99.995], rounded down, so the 99\% threshold is only just undetermined) and GLiNER Guard 548 (99.1\% [97.9; 99.7]). Fourth, the held-out sets come from the same generator on which the engine was developed: they protect against tuning to specific documents, not to the generator's style or to the hard formats the author chose.

On set~7, version 0.5 meets the four detection thresholds in both languages: identifiers, names, addresses and organisations, leaks and the strict count; in UK English it also meets both restoration thresholds. In Russian, restoration over all placeholders is undetermined and restoration of name placeholders is not met, mainly because pii-guard's inflection, which the engine uses to restore Russian names, produces wrong surname forms, and partly because some first names are stored wrongly (Section~\ref{sec:restore}). Its Russian strict count rose from 0.2\% on set~6 to 0.8\% [0.4; 1.6], mostly organisations; detection did not change between versions 0.4 and 0.5, so this is a difference between the documents of the two sets. Its UK leaks are two names; on noisy UK documents alone the upper end of the leak interval is 1.7\%. The restoration verdicts rest on the pre-registered rule, which bootstraps over answers when failures occur and uses the zero-failure bound otherwise; with rare failures the two can rank a cell with failures above a cell without, so we report answer-level rates next to it (Table~\ref{tab:rt7}).

\begin{table}[!t]
\centering\scriptsize\setlength{\tabcolsep}{3pt}
\begin{tabular}{>{\raggedright\arraybackslash}p{2.0cm}>{\raggedright\arraybackslash}p{2.2cm}>{\raggedright\arraybackslash}p{2.2cm}>{\raggedright\arraybackslash}p{1.8cm}>{\raggedright\arraybackslash}p{1.8cm}>{\raggedright\arraybackslash}p{2.5cm}>{\raggedright\arraybackslash}p{1.3cm}}
\toprule
Dataset (version) & Identifiers $\geq$99\% & Names, addr., org.\ $\geq$95\% & Leaks $\leq$2\% & Strict $\leq$2\%$^{c}$ & Restoration $\geq$99\%$^{d}$ & Precision \\
\midrule
RU hivetrace$^{a}$ & 427/427, $\geq$99.1\% & 404/404, $\geq$98.9\% & 0.0\% [0.0; 0.2] & 0.0\% [0.0; 0.2] & -- & 89.2\% \\
RU redmadrobot$^{a}$ & 98.7\% [98.0; 99.3] & 98.3\% [97.5; 99.0] & 1.1\% [0.7; 1.5] & 1.3\% [1.0; 1.8] & -- & 76.4\% \\
UK Nemotron$^{b}$ & 99.97\% [99.9; 100.0]$^{f}$ & 98.3\% [97.9; 98.8] \ok & 0.9\% [0.6; 1.3] \ok & 2.1\% [1.7; 2.7] & -- & 93.5\% \\
\quad original IDs only & 99.8\% [98.996; 99.995] \und & -- & -- & -- & -- & -- \\
RU set 4 (0.2) & all, $\geq$99.63\% & 99.8\% [99.7; 99.9] & 0.2\% [0.1; 0.7] & 2.5\% [1.7; 3.7] & 0 failures, 40 answers, $\geq$91.1\% & 98.5\% \\
UK set 4 (0.2) & all, $\geq$99.63\% & 98.7\% [98.5; 98.9] & 0.9\% [0.5; 1.7] & 15.5\% [13.4; 17.9] & 99.6\% [99.2; 99.9] & 99.8\% \\
RU set 5 (0.3) & all, $\geq$99.63\% & 99.7\% [99.6; 99.8] & 0.2\% [0.1; 0.7] & 3.3\% [2.4; 4.6] & 99.92\% [99.86; 99.97] (evaluator 1)$^{e}$; names 89.1--89.3\% (evaluator 3) & 98.5\% \\
UK set 5 (0.3) & all, $\geq$99.63\% & 98.9\% [98.7; 99.1] & 0.7\% [0.3; 1.4] & 13.8\% [11.8; 16.1] & 0 failures, 380 answers, $\geq$99.03\% & 99.8\% \\
RU set 6 (0.4) & all, $\geq$99.63\% & 99.98\% [99.96; 100.0] & 0.1\% [0.0; 0.6] & 0.2\% [0.1; 0.7] & 0 failures (evaluator 2)$^{e}$; names 93.8--95.0\% (evaluator 3) & 98.4\% \\
UK set 6 (0.4) & all, $\geq$99.63\% & 99.95\% [99.92; 99.99] & 0.6\% [0.3; 1.3] & 0.6\% [0.3; 1.3] & 0 failures, 380 answers, $\geq$99.03\% & 99.9\% \\
RU set 7 (0.5) & all, $\geq$99.63\% \ok & 99.93\% [99.88; 99.98] \ok & 0.3\% [0.1; 0.9] \ok & 0.8\% [0.4; 1.6] \ok & 98.86\% [98.64; 99.08] \und; names 94.58\% [93.56; 95.65] \no & 98.4\% \\
UK set 7 (0.5) & all, $\geq$99.63\% \ok & 99.98\% [99.96; 100.0] \ok & 0.2\% [0.1; 0.7] \ok & 0.2\% [0.1; 0.7] \ok & 99.99\% [99.98; 100.0] \ok; names 99.97\% [99.90; 100.0] \ok & 99.8\% \\
\bottomrule
\end{tabular}
\caption{pii-roundtrip on every dataset; precision is the share of masked fragments that touch personal data. Counts with $\geq$ are zero-failure results with exact one-sided 97.5\% lower bounds over the documents (or texts) that contain such values, rounded down; on held-out sets 4 to 7 no document has a missed identifier (17{,}489, 12{,}566, 17{,}298, 12{,}636, 17{,}291, 12{,}646, 17{,}245 and 12{,}636 identifiers). Verdicts are given only for set~7, the last pre-registered test, and for the UK external data; for set~7 the restoration cell gives all placeholders and name placeholders, each against 99\%. $^{a}$Home dataset of a component of the engine (GLiNER Guard and pii-guard respectively); reference only. $^{b}$Both subsets, 3{,}248 of 3{,}249 identifiers; in-distribution for GLiNER Guard. $^{c}$Pre-registered for sets~6 and~7 and applied only there. $^{d}$Model with the lower rate; with zero failures in both models, the bound for the number of answers. $^{e}$Overstated: the evaluator of that set could not see wrongly stored or wrongly inflected Russian names (Appendix~\ref{app:history}). $^{f}$No verdict: 83\% of these identifiers were injected by the generator that shares the engine's checksum code; the verdict is given on the original identifiers.}
\label{tab:engine}
\end{table}

\FloatBarrier
\section{Sequential testing and changes of rules}
\label{sec:sequential}
The engine was built in a loop of development, a test on a new held-out set, analysis of the errors and a new version (Table~\ref{tab:sets}). Each held-out result is an unbiased estimate for the version it tested, on documents of this generator and under the rules fixed before that set, because that version was frozen and run once; the rules for set~7 were themselves chosen after sets~5 and~6 had been seen. The external datasets were rescored after each change of metric; the engine was not changed in response to them after version 0.1, but its UK external verdicts are also the result of repeated looks. The loop as a whole is a sequence of tests with an implicit stopping rule (continue until a set passes), and every cycle fitted the engine more closely to the generator, which is an external-validity risk beyond the use of one generator. We therefore fixed in advance that set~7 is the last set for this paper and that its result is published whatever it is; set~7 did not pass every threshold. We make no correction for multiple cycles; inference is per set and per version.

Several rules of evaluation were also changed after data had been seen. Table~\ref{tab:rules} lists them with the direction in which each moved the engine's figures; the full list with dates and effects on the other tools is in the released \nolinkurl{results/rule_changes.md}. Changes in both directions occurred. The evaluator changes before set~6 made the engine look better on set~5, and the later ones (the stored-name check and evaluator version 3) made it look worse on sets~5 to~7.

\begin{table}[!htbp]
\centering\scriptsize\setlength{\tabcolsep}{3pt}
\begin{tabular}{>{\raggedright\arraybackslash}p{2.6cm}>{\raggedright\arraybackslash}p{1.3cm}>{\raggedright\arraybackslash}p{6.6cm}>{\raggedright\arraybackslash}p{3.4cm}}
\toprule
Date (release) & Data seen & Change & Effect on the engine \\
\midrule
19 Oct (0.2.1) & set 4 & Each issued placeholder occurrence scored once & up (fewer double counts) \\
19 Oct (0.2.2) & set 4 & Bracket-rewritten tags counted as issued and altered & none \\
19 Oct (0.2.2) & set 4 & Recall and precision by cluster bootstrap; ``borderline'' rule over seeds & wider intervals \\
19 Oct (0.2.2) & set 4 & Restoration threshold judged on clean answers & down \\
19 Oct (0.3.0, before set 5) & set 4 & Back to occurrences, bootstrap over answers; 380 answers for the engine & up \\
22 Oct (0.3.1) & sets 4, 5 & Leak metric counts dotted numbers & none (other tools worse) \\
22 Oct (0.3.1; threshold from set 6) & set 5 & Strict count of partly masked values & down on sets 4, 5 \\
22 Oct (0.4.0, before set 6) & set 5 & Evaluator version 2 (more case forms) & up; later found too lenient \\
23 Oct (0.4.0) & sets 5, 6 & Zero-failure bounds rounded down; bound over documents for recall & slightly down \\
23 Oct (0.4.1) & set 6 & Stored names checked against the annotation & down on sets 5, 6 \\
23 Oct (before set 7) & sets 5, 6 & Evaluator version 3; threshold for name placeholders & down \\
24 Oct (0.5.0) & set 7 & First names and patronymics of stored names also checked by hand & down (companion measure) \\
24 Oct (0.5.0) & set 7 & Manual check of all failures in three groups & none on the verdicts \\
\bottomrule
\end{tabular}
\caption{Changes to the rules of evaluation made after data had been seen (dates in October 2026). Two changes that affect only the other tools (privacy-filter listed without an organisation type; the restoration verdict of pii-guard on set~5 removed) and smaller presentational changes are in \nolinkurl{results/rule_changes.md}. The table starts with release 0.2.1; earlier corrections of the leak metric (Appendix~\ref{sec:review}) lowered the leak rates of every tool.}
\label{tab:rules}
\end{table}

\section{Limitations}
All data in this study are synthetic, and the engine was developed over six cycles on the generator of the held-out documents, so the results do not establish performance on real documents; the identifier results in particular are close to circular, and the UK external test is in-distribution for GLiNER Guard. The leak metric counts literal values only, not quasi-identifiers or attribute inference. Russian restoration by version 0.5 does not meet the thresholds: about one in twenty Russian name placeholders comes back with a wrong form of the name, mainly a wrong surname form (often a plural) produced by pii-guard's inflection partly first names the engine stores wrongly, and in a few cases its own proxy-name inflection; this is not fixed in the released version, so Russian restoration of name placeholders does not meet its threshold and restoration of all placeholders is undetermined (below 99\% under the companion measure). Evaluator version~3 cannot see a gender error where the masculine and feminine forms coincide. The restoration rule is not monotonic for rare failures, and we report answer-level rates next to it. The analyses added after the runs are exploratory: the leak breakdowns, the strict count on sets~4 and~5 and on external data, answer-level rates on sets~4 and~5, the format comparison, the manual checks of failures and the hand check of first names and patronymics of stored names on set~7. The other configurations' round trips were run on sets~4 and~5 only, with 20 documents per language (on set~5, 9 of them among the engine's 190). On short Russian texts the engine masks much more non-personal text than pii-guard (89.2\% of its masked fragments contain personal data on hivetrace, against 96.5\%); these are errors of the GLiNER component, which takes ordinary words and slang for personal data. The round trip uses two open models of 9--12 billion parameters at 4-bit quantisation on one computer; larger hosted models may treat placeholders differently, temperature~0 is not guaranteed to be deterministic across hardware, the restoration check does not verify that each value returns to the right place, and each evaluator version has known errors on Russian names (version~3 counted 8 correct restorations as failures on set~7). Models invent placeholders that no tool can restore and that remain visible to the user; the restoration rate does not include them. Utility of the answers is measured only by whether they keep placeholders. The redmadrobot dataset contains many identifiers with invalid checksums (218 of 223 SNILS numbers are invalid or written in words) and numbers written in words, and Nemotron-PII contains annotation noise. We did not measure promptShield or commercial services, and the round trip has no prompt control for Prompt Anonymizer or the engine. The benchmark was designed, run and analysed by the author of the engine.

\section{Conclusion}
On external datasets, the best open tools for Russian and UK English do not meet all predefined thresholds; their identifier recall is 96--98\% against a 99\% threshold, borderline in Russian and not met in UK English. On synthetic documents the best existing tools leak in more than half of the documents, in 38--41\% of those without added noise, and pii-guard in 32\% if organisations, which it does not support, are ignored (both post-hoc breakdowns). A thin layer of checksum rules, whole-value masking of addresses and organisations, document-level propagation, entity resolution and tolerant restoration on top of existing open models reduces document leaks on the last held-out set, registered publicly before the run with a stopping rule and drawn from the generator on which the engine was developed, from 60.8\% [57.7; 63.8] to 0.3\% [0.1; 0.9] for Russian and from 58.3\% [55.2; 61.3] to 0.2\% [0.1; 0.7] for UK English; any part of a value stays visible in 0.8\% and 0.2\% of documents, and 98.4\% and 99.8\% of the masked fragments contain personal data. In UK English the engine also restores at least 99\% of its placeholders, names included, after the answers of two local models. The existing tools were run with documented defaults and were not tuned to this generator. In Russian the engine does not restore reliably: 94.6\% of name placeholders come back correctly, mainly because the inflection it borrows from pii-guard produces wrong surname forms, partly because it stores some first names wrongly, and in a few cases because of its own proxy-name inflection. Restoring Russian names without that inflection, and testing it on new data, is the main open problem. Whether these results hold on real documents is not established here.

\section*{Declarations}
\paragraph{Competing interests.} The author develops pii-roundtrip, the engine evaluated here, and operates Besk Tech, under whose name the engine is published. pii-roundtrip is open source under the Apache 2.0 licence, and its desktop application for macOS is distributed free of charge. The study was designed to decide whether a product of this kind is worth building. The project brief, in a section amended on 13 October 2026, defines the market signal for that decision as the response to the published open engine and benchmark during three months after publication; a paid application would be considered only if at least five potential users ask for a ready product in that time. No commercial product exists at the time of writing. The author has no relationship with the developers of the other tools evaluated here.

\paragraph{Funding.} This research received no external funding.

\paragraph{Use of AI tools.} Claude Opus 5.5 (Anthropic, model identifier claude-opus-5-5), through Claude Code and agent sessions, was used to write and refactor code, to run the benchmark, to draft and edit the text of this paper, and to carry out the code and paper reviews described in Appendix~\ref{sec:review}, which were run as separate sessions without access to the development sessions. The author designed the study, set the thresholds, checked the results and takes full responsibility for the content. All numbers in the paper are produced by the released scripts from the released predictions and answers, except those of the manual checks (classification of round-trip failures and of stored-name mismatches), whose item-by-item verdicts are released.

\paragraph{Data and code availability.} The engine, the benchmark and evaluation code, the generator, held-out sets 1 to 7 with their SHA-256 hashes, the pre-registrations of sets~5 to~7 (\texttt{HOLDOUT.md}, commits 8b8e933, 444c4bd and 13c42b8), the per-document predictions of every tool, the raw LLM answers of the round trips, the run log, the model versions and the scripts that produce every number in this paper (including \texttt{bench/eval/cluster\_ci.py}, \texttt{bench/eval/leak\_breakdown.py} and \texttt{bench/eval/cycle.py}) are available at \url{https://github.com/beskvladimir-create/pii-roundtrip} (release \RELVERSION) under the Apache 2.0 licence and archived on Zenodo (DOI: \href{https://doi.org/\RELDOI}{\RELDOI}). Intervals for recall and precision come from \nolinkurl{results/cluster_ci.json}, leak breakdowns from \nolinkurl{results/leak_breakdown.holdout7.json}, the strict count from \nolinkurl{results/strict_leaks.json}, round-trip results from \nolinkurl{results/cycle_metrics.holdout7.json} and \nolinkurl{results/cycle_extras.holdout7.json}, evaluator versions from \nolinkurl{results/evaluator_versions.holdout7.json}, stored names from \nolinkurl{results/stored_names.holdout7.json} and \nolinkurl{results/stored_names_review.holdout7.json}, the manual check of failures from \nolinkurl{results/failures_review.holdout7.json}, the rule changes from \nolinkurl{results/rule_changes.md}, and the run journals from \nolinkurl{results/holdout5_cycle_journal.md}, \nolinkurl{results/holdout6_cycle_journal.md} and \nolinkurl{results/holdout7_cycle_journal.md}; the corresponding files for earlier sets are in the same folder. The development set is not released, and no number in this paper depends on it. External datasets are downloaded at pinned revisions and are not redistributed; the released predictions on them contain only character offsets, and short excerpts appear only as error examples in the results and the report.

\appendix
\section{Development history and earlier held-out sets}
\label{app:history}
This appendix describes sets~1 to~6 in order and then the engine's round trips on sets~4 to~6.

Sets~1 and~2 (100 documents per language each) tested version 0.1 of the engine. Set~3 (1{,}000 documents per language) tested a candidate of version 0.2: it leaked in 0.5\% [0.2; 1.2] of Russian and 3.7\% [2.7; 5.1] of UK documents, so it failed the UK leak threshold. Its errors were then analysed and fixed, so sets~1 to~3 are development data. Set~4 (1{,}000 documents per language) was generated and its SHA-256 hash recorded before that analysis, and each tool was run on it once on 17 October 2026 (UTC); the hashes and the run log were kept in the private development history, so the order of events for sets~1 to~4 rests on that log and not on an external timestamp. Set~4 tested version 0.2: the engine leaked in 0.2\% [0.1; 0.7] of Russian and 0.9\% [0.5; 1.7] of UK documents, and its round trip showed that its restorer did not recognise placeholders whose brackets the model had dropped (see the last paragraph of this appendix). Version 0.3 changes only the restorer; the engine's detection code is unchanged since version 0.2.0, so detection results on the external datasets and on set~4 apply to version 0.3 as well.

Set~5 (1{,}000 documents per language, generator seed 20261020) tested version 0.3. Its files, their SHA-256 hashes, the decision rules and the run plan were published in the public repository (commit 8b8e933, 19 October 2026, 17:49 UTC) before any tool was run on it; the runs began on 20 October 2026. The engine's round trip was interrupted once, during its 72nd Russian answer of gemma4:12b, when another session on the same machine stopped the Ollama server; that session had also sent its own requests to the server for a few minutes before the 71st answer was saved. The round trip was resumed from the saved answers; the run log records the detection runs, and the round trip is reconstructed in the released \nolinkurl{results/holdout5_cycle_journal.md}. On set~5 version 0.3 met the four thresholds of the brief, but a review of this paper found that it left part of an address or organisation visible in 3.3\% of Russian and 13.8\% of UK documents, and that some of its Russian restoration failures were real (see the last paragraph of this appendix). Its errors were analysed, so sets~1 to~5 are development data for version 0.4. The pre-registration file of set~5 carries dated corrections of 22 October 2026 (its generation date and the number of sets), below the unchanged original text.

Set~6 (1{,}000 documents per language, generator seed 20261022; 17{,}291 Russian and 12{,}646 UK structured identifiers, 12{,}966 Russian and 13{,}164 UK names, addresses and organisations) tested version 0.4. It was generated after the changes of version 0.4 with the unchanged generator, and its files, hashes, the code under test (with the git tree hashes of the engine and the benchmark code), the decision rules and the run plan were published in the public repository (commit 444c4bd, pushed on 22 October 2026 at 07:57 UTC) before any tool was run on it. The round trip was stopped once during a mains power cut and continued by the same command from the saved answers (\nolinkurl{results/holdout6_cycle_journal.md}). The evaluator did not change between the set-6 pre-registration and release 0.4.0. The results of every tool on sets~4 to~6 are released.

\paragraph{Engine round trips on sets 4 to 6.} On set~4, version 0.2 lost 6 of 1{,}455 UK occurrences with gemma4:12b and 1 of 1{,}201 with qwen3.5:9b, because its restorer did not recognise placeholders whose brackets the model had dropped (\texttt{[ORG\_1}, \texttt{ORG\_1]}); version 0.3 restores these forms, and on set~5 it restored every UK placeholder over 380 answers per model although gemma4:12b altered 42 and qwen3.5:9b 5 of them. In Russian on set~5, 10 occurrences in 9 answers (qwen3.5:9b) and 8 in 6 answers (gemma4:12b) failed under evaluator version 1; a manual check found 11 of the 18 to be correct restorations that version 1 rejected, one a name in the wrong case, and six real (two where pii-guard's inflection dropped the surname of the second of two adjacent name placeholders, four where a first name that was not declined was not stored). Both were fixed in version 0.4. On set~6, version 0.4 lost no occurrence under the pre-registered evaluator version 2 in 1{,}520 answers. A manual check of the version-1 failures, and then a check of every stored Russian name against the annotation (release 0.4.1), found that 151 of 1{,}097 stored Russian names were wrong (surnames in a plural form such as \emph{Abramy} for \emph{Abramov}, wrong gender or case), which version 2 could not see because it compares the answer with the stored value. Rescored with evaluator version 3 after the fact, set~6 gives 93.8--95.0\% for Russian name placeholders and 98.7\% for all Russian placeholders with gemma4:12b; on set~5 (version 0.3) name placeholders give 89.1--89.3\%. This led to version 0.5 and to set~7 with evaluator version 3 and a separate threshold for name placeholders. The pre-registered restoration verdicts of sets~5 and~6 therefore overstate Russian restoration.

\section{Deviations from the brief}
\label{app:deviations}
The study deviates from the brief as follows.
\begin{enumerate}\setlength{\itemsep}{0pt}
\item The brief foresaw one held-out set run once at the end; held-out sets~1 to~3 were run and then used for development, set~4 tested version 0.2 of the engine and led to a change of its restorer, set~5 tested version 0.3 and led, through the review of this paper, to version 0.4, set~6 tested version 0.4 and led to version 0.5, and set~7 tests the released version 0.5 (Section~\ref{sec:heldout}).
\item The brief excluded the external datasets from development; two changes of engine version 0.1 followed an analysis of misses on the Russian external datasets.
\item The brief specified Wilson intervals; recall and precision now use a document-level cluster bootstrap.
\item The brief listed a local LLM as a detector; it was run during development but is not reported.
\item The brief specified Presidio for UK English with a GLiNER recogniser; we ran it with spaCy recognisers only. For Russian the brief listed a recogniser for the compulsory health insurance (OMS) policy number; none was written, and OMS numbers are not part of the data.
\item The brief asked for exact-boundary and partial-overlap matching and per-type results; these are computed and released but not reported here.
\item The brief asked to report restoration errors on verbatim and model-altered placeholders separately; we report them together and give the number of altered placeholders and of failures among them.
\item The data grew from 100 to 1{,}000 documents per language.
\item The brief foresaw a measurement of contextual re-identification and a manual test of promptShield; neither was carried out.
\item The brief foresaw sending the results to the authors of pii-guard and GLiNER Guard one to two weeks before publication and contributing fixes to their projects; neither was done.
\item The brief foresaw the round trip on the documents of the first stage; it was run on held-out sets~4 and~5 for all configurations and on sets~6 and~7 for the engine only.
\item The round-trip controls (generic prompt, tolerant parser) were added after a review of an earlier, unpublished draft of this paper.
\item The brief took the decision on the engine, too, on the external datasets; since both Russian external datasets are home datasets of the engine's components and no external data exist for the round trip, the engine's verdicts rest on the held-out synthetic sets.
\item The brief asked to configure each tool for its best result; we used each tool's documented defaults and did not tune thresholds.
\item The brief specified Wilson intervals for restoration as well; we bootstrap over answers.
\item The brief required one placeholder per person; ambiguous mentions now receive their own placeholder.
\item The unit of the restoration threshold changed twice: occurrences until release 0.2.1, answers in release 0.2.2 (19 October 2026, with Wilson intervals over answers, after the set-4 round trip was known), and occurrences again in the pre-registration of set~5 the same evening. The brief speaks of placeholders, which supports the occurrence unit; but under the answer rule of 0.2.2 even 380 answers without failure give a lower bound of 98.99\%, which cannot clear the threshold, and the Russian result on set~5 would be not met (371 of 380 answers without failure for qwen3.5:9b, Wilson [95.6; 98.7]).
\item The leak metric was corrected after the set-5 run (Section~\ref{sec:metrics}). The brief was amended on 13 October 2026, after the first results, in its market-signal and second-stage sections, which are marked as amended; the threshold section was not marked as changed. The SHA-256 hash of the brief at the first commit is recorded in the pre-registration file; the brief itself is not released. As the brief required, Nemotron texts in which an original number also occurs outside the annotation were excluded.
\item The strict count of documents with any value not fully masked is not in the brief; it was added after review as an exploratory measure on set~5 and pre-registered with a 2\% threshold for sets~6 and~7. Its code counted every annotated type until release 0.3.1 and only the threshold types from release 0.4.0; the set-5 figures did not change, a few set-4 and external figures did.
\item The restoration evaluator was revised before set~6 (version 2) and before set~7 (version 3, Section~\ref{sec:metrics}); each set is reported with the version fixed before it, and sets~5 to~7 are rescored with all versions in the release.
\item The check of stored Russian names was added after the set-6 run and pre-registered as a companion measure for set~7; with every mismatch to be checked by hand. Comparing first names and patronymics as well, and grouping the failures into three kinds, was not pre-registered.
\item Evaluator version~3 and a sixth threshold, restoration of at least 99\% of name placeholders, were fixed before set~7; neither is in the brief.
\item The pre-registration of set~7 added a stopping rule: set~7 is the last set for this paper, and its result is published whatever it is.
\end{enumerate}

\section{What AI-run reviews and checks changed}
\label{sec:review}

Two code reviews by separate AI sessions without access to the development sessions, one of the evaluation code and one of the engine, found errors that changed conclusions of an interim version of this work (version 0.1); our own checks and further AI reviews of this paper found more. We report them because similar errors are easy to make in PII benchmarks.

\paragraph{Leak detection.} To match numbers across separators, the leak detector joined adjacent alphanumeric tokens, which glued an English word to the following number (``NHS number 943 476 5919'' became ``number9434765919'') and hid the leak. After the fix, GLiNER Guard's UK leak rate rose from 2.5\% to 4.0\% in the interim version. A later check found the opposite error: a first name that is also an ordinary word of the text (``June'') counted as a leak even when the whole name was masked.

\paragraph{Clustered errors.} The interim version reported that pii-guard fails restoration in Russian at 95.9\%. Nineteen of 24 errors came from a single answer; with intervals bootstrapped over answers instead of placeholders the result was undetermined. A review of this paper found that recall and precision intervals had the same problem, so they are now bootstrapped over documents; this changed the verdict for pii-guard's Russian identifier recall from not met to undetermined.

\paragraph{Restoration metric.} The restoration metric added placeholders left in the text to lost values, so a placeholder that stayed in the text and whose value was therefore missing counted twice. Each issued occurrence is now scored once. The unit of the threshold then moved to answers and back to occurrences (Appendix~\ref{app:deviations}, item on the restoration unit). A later review found that the evaluator did not count placeholders that a model had rewritten in square brackets (\texttt{[PII type="PASSPORT" id="1"]}, \texttt{[Name\_2]}): 43 such tags in one pii-guard answer and 29 in one Prompt Anonymizer answer on set~4 were neither counted nor scored, and the two cells had been reported as 99.8\% and 100\%. The evaluator now counts them as issued and altered, and the two cells fall to 96.8\%.

\paragraph{Format or parser.} An earlier draft concluded that restoration failures were caused by the placeholder format. A review pointed out that format, system prompt and restorer were confounded, and a later review that the tolerant parser restores by construction. With the controls of Appendix~\ref{app:other} and a second held-out set, the difference in alteration rates between formats is not established; what holds is that the parser decides whether an alteration becomes a loss, and that a prompt can make a model drop placeholders altogether.

\paragraph{Dropped brackets and a new held-out set.} The round trip on set~4 showed that the engine's own restorer did not recognise placeholders with a dropped bracket. Fixing this and reporting the fixed engine on set~4 would have tuned it to its own test, so version 0.3 was tested on set~5, published with its decision rules and run plan before the run.

\paragraph{Partly masked values and stored names.} A review of version 0.3 on set~5 showed that the pre-registered leak metric hid addresses and organisations masked only in part (Section~\ref{sec:metrics} and Appendix~\ref{app:history}). Version 0.4 masks them whole and was tested on set~6, with the stricter count pre-registered as a fifth threshold. After the set-6 run, a manual check of the evaluator's version-1 failures led to the discovery that the engine stores some Russian surnames in a wrong form; the pre-registered evaluator and value-fidelity measure could not see this, and we report it as an error of version 0.4; version 0.5 fixed the storage of most surnames, but 30 of 1{,}027 Russian names, mostly first names, are still stored wrongly on set~7 (Section~\ref{sec:restore}).

\paragraph{Circular and in-distribution evidence.} The claim that checksum rules find all NHS numbers rested on numbers and rules derived from the same generator. It holds for well-formed numbers but is not evidence about real documents. A review also found that the share of original UK identifiers had been understated (553, not 59) and that GLiNER Guard was trained partly on Nemotron-PII.

\paragraph{Entity merging.} In the engine, people sharing a surname received the same placeholder: a landlord and a tenant became one party, and restoration could name the wrong person. This failure is worse than a leak because a user is unlikely to notice it, and our restoration metric would not detect it. Resolution now uses all name parts and gender and does not merge ambiguous mentions.

\paragraph{Asymmetric comparison and reproducibility.} Version 0.1 compared a competitor without its home dataset with the engine on all datasets; comparisons now use identical data, and the engine's results on both Russian home datasets are reported without verdicts. The synthetic generator depended on the date of the run through Faker, so regenerating a held-out set on another day differed from the frozen file in one e-mail address; the date is now fixed. The round-trip cache keyed LLM answers by document rather than by prompt, which would have reused answers to an older anonymisation; it is now keyed by a hash of the prompt.

\section{Round trip of the other configurations}
\label{app:other}
Table~\ref{tab:rt} reports the round trip on set~5 for every reversible configuration and on set~6 for the engine. A configuration either uses the tool's own reversible format (pii-guard's XML-style tags such as \texttt{<PII type="INN" id="4" />}, Prompt Anonymizer's \texttt{<Name\_1>}) or wraps a detector in plain bracketed placeholders such as \texttt{[PERSON\_1]} through the benchmark's own wrapper, which also has its own system prompt with examples. To separate format, prompt and parser, pii-guard's tags and the plain wrappers were also run with a short generic prompt (there is no such control for Prompt Anonymizer or the engine), and pii-guard's answers were restored a second time (``re-parsed tolerantly'') with a tolerant parser that recognises its tags by type and number. Every cell of the round trip was also run on set~4 with 40 answers, there on the same 20 documents per language for all configurations; those results are released, and we refer to them where they differ. On set~5 the other configurations used 20 documents per language, 9 of which are among the engine's 190, so set-5 comparisons between the engine and the other tools are on different documents. Values that leaked into the anonymised input could count as restored; excluding them changes no failure count in any cell on either set.

\begin{table}[!t]
\centering\scriptsize\setlength{\tabcolsep}{2pt}
\setlength{\tabcolsep}{1.5pt}
\begin{tabular}{lllll}
\toprule
 & \multicolumn{2}{c}{qwen3.5:9b} & \multicolumn{2}{c}{gemma4:12b} \\
Configuration & ans.\ / occ.\ / fail. & restoration & ans.\ / occ.\ / fail. & restoration \\
\midrule
\multicolumn{5}{l}{\emph{Russian}} \\
pii-guard tags, own & 15/40 / 542 / 1 & 99.8\% [99.4; 100.0] & 40/40 / 1{,}564 / 0 & 100\% ($\geq$91.1\%) \\
\quad re-parsed tolerantly & 15/40 / 542 / 0 & 100\% ($\geq$78.1\%) & 40/40 / 1{,}564 / 0 & 100\% ($\geq$91.1\%) \\
pii-guard tags, generic & 26/40 / 1{,}060 / 3 & 99.7\% [99.2; 100.0] & 40/40 / 1{,}703 / 29 & 98.3\% [94.6; 100.0] \\
\quad re-parsed tolerantly & 26/40 / 1{,}060 / 0 & 100\% ($\geq$86.7\%) & 40/40 / 1{,}703 / 0 & 100\% ($\geq$91.1\%) \\
pii-guard + plain, wrapper & 40/40 / 1{,}961 / 0 & 100\% ($\geq$91.1\%) & 40/40 / 2{,}168 / 0 & 100\% ($\geq$91.1\%) \\
pii-guard + plain, generic & 40/40 / 1{,}956 / 0 & 100\% ($\geq$91.1\%) & 40/40 / 2{,}297 / 0 & 100\% ($\geq$91.1\%) \\
pii-roundtrip 0.3, set 5$^{a}$ & 380/380 / 12{,}274 / 10 & 99.92\% [99.86; 99.97] & 380/380 / 13{,}722 / 8 & 99.94\% [99.88; 99.99] \\
pii-roundtrip 0.4, set 6$^{a}$ & 380/380 / 12{,}720 / 0 & 100\% ($\geq$99.03\%)$^{b}$ & 380/380 / 13{,}481 / 0 & 100\% ($\geq$99.03\%)$^{b}$ \\
\midrule
\multicolumn{5}{l}{\emph{UK English}} \\
Prompt Anonymizer tags & 40/40 / 923 / 0 & 100\% ($\geq$91.1\%) & 40/40 / 1{,}132 / 0 & 100\% ($\geq$91.1\%) \\
GLiNER + plain, wrapper & 40/40 / 1{,}644 / 0 & 100\% ($\geq$91.1\%) & 40/40 / 1{,}923 / 3 & 99.8\% [99.6; 100.0] \\
GLiNER + plain, generic & 40/40 / 1{,}595 / 0 & 100\% ($\geq$91.1\%) & 40/40 / 1{,}900 / 2 & 99.9\% [99.7; 100.0] \\
pii-roundtrip 0.3, set 5$^{a}$ & 380/380 / 11{,}963 / 0 & 100\% ($\geq$99.03\%) & 380/380 / 13{,}857 / 0 & 100\% ($\geq$99.03\%) \\
pii-roundtrip 0.4, set 6$^{a}$ & 380/380 / 11{,}792 / 0 & 100\% ($\geq$99.03\%) & 380/380 / 13{,}803 / 0 & 100\% ($\geq$99.03\%) \\
\bottomrule
\end{tabular}
\caption{Round trip on held-out set~5, and on set~6 for engine version 0.4 (summary and reply; 40 answers per cell; $^{a}$380 answers; set~5 scored with evaluator version 1, set~6 with version 2). $^{b}$Overstated: some Russian names were stored in a wrong form, which the evaluator cannot see (Appendix~\ref{app:history}); this also applies to the Russian rows of engine 0.3 on set~5, whose name placeholders score 89.1--89.3\% with evaluator version 3. ``Own'', ``wrapper'' and ``generic'' name the system prompt. ``Ans.'' counts answers with placeholders, ``occ.'' tool-issued placeholder occurrences in the answers and ``fail.'' those that failed. Intervals: bootstrap over answers; an upper bound of 100.0 is the bootstrap's ceiling. Values in parentheses are the pre-registered lower bounds for zero failures, $0.025^{1/n}$ with $n$ the answers with placeholders.}
\label{tab:rt}
\end{table}

\paragraph{pii-guard's tags.} With pii-guard's own parser, its tags lost values under both prompts. With its own prompt, qwen3.5:9b used no tag at all in 25 of 40 answers (27 of 40 on set~4), describing the data in words or omitting it, which no parser can repair. With the generic prompt, gemma4:12b wrote 29 tags in one answer as opening tags without the closing slash (\texttt{<PII type="PERSON" gender="male" id="1">}), and the strict parser lost all 29; on set~4 the same model did this with 44 tags in two answers. The tolerant parser, written after the set-4 failures but fixed before set~5, restored every counted tag on both sets. It inserts the stored value without re-inflecting Russian names, which passes the evaluator's check trivially but can produce ungrammatical Russian. It recognises tags with the same pattern that the evaluator uses to count them, so it restores every counted tag by construction; its rows show only that no corrupted tag lost its type or number. The remaining failures of pii-guard's own parser on set~5 (one and three occurrences with qwen3.5:9b) concern one name in one document (ru-medical-195) and are real: pii-guard stored the noisy name in an altered form and restored it in a third form; they do not occur with the tolerant parser, which inserts the stored value as it is.

\paragraph{Placeholder format.} On set~4, after the evaluator was corrected to count tags rewritten in square brackets (Appendix~\ref{sec:review}), models altered pii-guard's tags in 101 of 5{,}099 occurrences, in 9 of 127 answers with tags, against 1 of 8{,}099 plain placeholders, in 1 of 160 answers, around the same detector. On set~5 the counts were 29 of 4{,}869 occurrences in 1 of 121 answers against 4 of 8{,}382 in 1 of 160 answers. Alterations concentrate in a few answers, so occurrence counts overstate the evidence, and at the level of answers the two sets disagree. We therefore do not claim that models alter XML-style tags more often than plain placeholders. What holds on both sets is that pii-guard's strict parser turns almost every altered tag into a loss (99 of 101 on set~4, 29 of 29 on set~5); on set~4 three intact tags failed as well, at least one because the parser's clean-up removed it together with a neighbouring altered tag. A parser that accepts altered forms restores them.

\paragraph{Plain placeholders.} The benchmark's plain wrapper lost no occurrence around pii-guard. Around GLiNER Guard, gemma4:12b lost 3 and 2 occurrences under the wrapper and the generic prompt, where the model added words to the placeholder (\texttt{[RECIPIENT\_ADDRESS\_1]} and \texttt{[ADDRESS\_OF\_ORG\_1]}; \texttt{[YOUR\_ADDRESS\_1]}) that the wrapper does not strip. Prompt Anonymizer lost nothing on set~5; on set~4, qwen3.5:9b wrote 29 of its tags in one answer in square brackets (\texttt{[Name\_2]}), and all 29 were lost, and gemma4:12b lost 1 of 1{,}108.


\begin{thebibliography}{99}
\raggedright
\bibitem{gliner} U.~Zaratiana, N.~Tomeh, P.~Holat, T.~Charnois. GLiNER: Generalist Model for Named Entity Recognition using Bidirectional Transformer. In \emph{Proceedings of NAACL 2024}, pages 5364--5376. doi:10.18653/v1/2024.naacl-long.300.
\bibitem{gliner2} U.~Zaratiana, G.~Pasternak, O.~Boyd, G.~Hurn-Maloney, A.~Lewis. GLiNER2: Schema-Driven Multi-Task Learning for Structured Information Extraction. In \emph{Proceedings of EMNLP 2025: System Demonstrations}, pages 130--140. doi:10.18653/v1/2025.emnlp-demos.10.
\bibitem{glinerguard} B.~Minko, S.~Sadiekh, E.~Kokuykin. GLiNER Guard: Unified Encoder Family for Production LLM Safety and Privacy. arXiv:2605.05277, 2026.
\bibitem{piiguard} redmadrobot. pii-guard: Russian-language PII detection, anonymisation and reversible pseudonymisation. \url{https://github.com/redmadrobot-rnd/pii-guard}, commit 24230ab, 2026.
\bibitem{presidio} Presidio: data protection and de-identification SDK. \url{https://github.com/data-privacy-stack/presidio}, version 2.2.364.
\bibitem{opf} OpenAI. Privacy Filter. \url{https://github.com/openai/privacy-filter}, commit f7f00ca, 2026.
\bibitem{promptanon} akazah. Prompt Anonymizer. \url{https://github.com/akazah/prompt-anonymizer}, version 0.3.3, 2026.
\bibitem{nemotron} NVIDIA. Nemotron-PII dataset. \url{https://huggingface.co/datasets/nvidia/Nemotron-PII}, CC BY 4.0.
\bibitem{lison} P.~Lison, I.~Pil\'an, D.~S\'anchez, M.~Batet, L.~{\O}vrelid. Anonymisation Models for Text Data: State of the art, Challenges and Future Directions. In \emph{Proceedings of ACL-IJCNLP 2021}, pages 4188--4203. doi:10.18653/v1/2021.acl-long.323.
\bibitem{tab} I.~Pil\'an, P.~Lison, L.~{\O}vrelid, A.~Papadopoulou, D.~S\'anchez, M.~Batet. The Text Anonymization Benchmark (TAB): A Dedicated Corpus and Evaluation Framework for Text Anonymization. \emph{Computational Linguistics} 48(4):1053--1101, 2022. doi:10.1162/coli\_a\_00458.
\bibitem{ratbench} N.~Kr\v{c}o, Z.~Yao, M.~Meeus, Y.-A.~de~Montjoye. RAT-Bench: A Comprehensive Benchmark for Text Anonymization. arXiv:2602.12806, 2026.
\bibitem{metricsurvey} Y.~Ren, K.~Ramesh, Y.~Yao, A.~Field. How do we measure privacy in text? A survey of text anonymization metrics. In \emph{Findings of IJCNLP-AACL 2025}, pages 1532--1544. doi:10.18653/v1/2025.findings-ijcnlp.94.
\bibitem{beyondmem} R.~Staab, M.~Vero, M.~Balunovi\'c, M.~Vechev. Beyond Memorization: Violating Privacy via Inference with Large Language Models. In \emph{Proceedings of ICLR 2024}. arXiv:2310.07298.
\bibitem{advanon} R.~Staab, M.~Vero, M.~Balunovi\'c, M.~Vechev. Language Models are Advanced Anonymizers. In \emph{Proceedings of ICLR 2025}. arXiv:2402.13846.
\bibitem{has} Y.~Chen, T.~Li, H.~Liu, Y.~Yu. Hide and Seek (HaS): A Lightweight Framework for Prompt Privacy Protection. arXiv:2309.03057, 2023.
\bibitem{genpseudo} S.~Hou, R.~Shang, Z.~Long, X.~Fu, Y.~Chen. A General Pseudonymization Framework for Cloud-Based LLMs: Replacing Privacy Information in Controlled Text Generation. arXiv:2502.15233, 2025.
\bibitem{privacyrestore} Z.~Zeng, J.~Wang, J.~Yang, Z.~Lu, H.~Li, H.~Zhuang, C.~Chen. PrivacyRestore: Privacy-Preserving Inference in Large Language Models via Privacy Removal and Restoration. In \emph{Proceedings of ACL 2025}, pages 10821--10855. doi:10.18653/v1/2025.acl-long.532.
\bibitem{papillon} S.~Li, V.~C.~Raghuram, O.~Khattab, J.~Hirschberg, Z.~Yu. PAPILLON: Privacy Preservation from Internet-based and Local Language Model Ensembles. In \emph{Proceedings of NAACL 2025}, pages 3371--3390. doi:10.18653/v1/2025.naacl-long.173.
\bibitem{promptgraph} C.~Gu, H.~Wan, D.~Hu et al. PromptGraph: Graph-Guided Prompt Sanitization for Balancing Privacy and Utility in LLM Inference. arXiv:2607.10709, 2026.
\bibitem{drsl} W.~Hu, Y.~Yu, X.~Wang. Reversibility-Verified De-identification for Cloud-Local LLM Inference: A Locally Certified Dehydrate-Rehydrate Loop with Layered Assurance (DR-SL). arXiv:2609.14883, 2026.
\bibitem{memprivacy} Y.~Chen, J.~Zhao, B.~Tang et al. MemPrivacy: Privacy-Preserving Personalized Memory Management for Edge-Cloud Agents. arXiv:2605.09530, 2026.
\bibitem{camp} A.~Panjwani. CAMP: Cumulative Agentic Masking and Pruning for Privacy Protection in Multi-Turn LLM Conversations. arXiv:2604.16521, 2026.
\bibitem{anonimpact} T.~Deu\ss{}er, M.~Hahnb\"uck, L.~Sparrenberg et al. On the Impact of Anonymization on the Performance of Large Language Models. arXiv:2609.11335, 2026.
\end{thebibliography}
\end{document}